\documentclass[lettersize,journal]{IEEEtran}
\usepackage{amsmath,amsfonts}
\usepackage{algorithmic}
\usepackage{algorithm}
\usepackage{array}
\usepackage[caption=false,font=normalsize,labelfont=sf,textfont=sf]{subfig}
\usepackage{textcomp}
\usepackage{stfloats}
\usepackage{url}
\usepackage{verbatim}
\usepackage{graphicx}
\usepackage{cite}
\usepackage{multirow}
\usepackage{hyperref}
\begin{document}

\title{Real-Time LPV-Based Non-Linear Model Predictive Control for Robust Trajectory Tracking in Autonomous Vehicles}

\author{Nitish Kumar,~\IEEEmembership{Graduate Student Member, IEEE,} Rajalakshmi Pachamuthu,~\IEEEmembership{Senior Member, IEEE}
\thanks{Nitish Kumar and Rajalakshmi Pachamuthu are with the Department
of Electrical Engineering, Indian Institute of Technology Hyderabad
(IITH), Sangareddy 502284, India (e-mail: ee22mtech02005@iith.ac.in;
raji@ee.iith.ac.in)}
}
\markboth{IEEE TRANSACTIONS ON INTELLIGENT VEHICLES}%
{Shell \MakeLowercase{\textit{et al.}}: A Sample Article Using IEEEtran.cls for IEEE Journals}

\maketitle

\begin{abstract}
This paper presents the development and implementation of a Model Predictive Control (MPC) framework for trajectory tracking in autonomous vehicles under diverse driving conditions. The proposed approach incorporates a modular architecture that integrates state estimation, vehicle dynamics modeling, and optimization to ensure real-time performance. The state-space equations are formulated in a Linear Parameter Varying (LPV) form, and a curvature-based tuning method is introduced to optimize weight matrices for varying trajectories. The MPC framework is implemented using the Robot Operating System (ROS) for parallel execution of state estimation and control optimization, ensuring scalability and minimal latency. Extensive simulations and real-time experiments were conducted on multiple predefined trajectories, demonstrating high accuracy with minimal cross-track and orientation errors, even under aggressive maneuvers and high-speed conditions. The results highlight the robustness and adaptability of the proposed system, achieving seamless alignment between simulated and real-world performance. This work lays the foundation for dynamic weight tuning and integration into cooperative autonomous navigation systems, paving the way for enhanced safety and efficiency in autonomous driving applications.
\end{abstract}


\section{Introduction}
\IEEEPARstart{A}{utonomous} vehicles (AVs) have revolutionized the transportation industry by offering safer, more efficient, and intelligent mobility solutions. A crucial aspect of autonomous driving is path tracking, which ensures that the vehicle follows a desired trajectory accurately and smoothly under varying driving conditions. Effective path tracking not only requires precise control of vehicle dynamics but also demands real-time computational efficiency to adapt to the constantly changing environment. In this regard, Model Predictive Control (MPC) \cite{ref139} \cite{ref140} has emerged as a prominent approach due to its predictive capabilities, optimization-based formulation, and ability to handle constraints on system states and control inputs.

While MPC has been successfully applied for path tracking,\cite{ref138} \cite{ref141} \cite{l1} existing solutions often fall short of achieving real-time performance when integrated with multiple sensors and complex vehicle dynamics. Traditional approaches typically assume fixed speed control with minimal adaptability, which does not reflect the dynamic behavior observed in manual driving \cite{l2} \cite{2_39}. In manual mode, human drivers intuitively adjust the vehicle's speed based on environmental factors, road conditions, and personal comfort preferences. Emulating this behavior in autonomous systems introduces significant computational challenges, particularly when real-time sensor integration, \cite{ref_2020} next-state prediction, and optimization need to be performed simultaneously.

To address these challenges, this paper presents a novel MPC-based path tracking framework capable of operating in two distinct speed modes—fixed-speed mode and variable-speed mode—that replicate the flexibility of manual driving. In the proposed framework: \begin{itemize} \item The fixed-speed mode maintains a constant speed along the desired trajectory, ensuring stability and predictability under uniform road conditions. \item The variable-speed mode dynamically adjusts the vehicle speed in response to external factors such as sharp turns, path curvature, and real-time sensor inputs, mimicking human-like driving behavior. \end{itemize}

To achieve real-time computational efficiency, a key challenge in implementing MPC, this work leverages parallelized computations to optimize the controller's performance. By dividing tasks such as next-state prediction, cost function evaluation, and optimization into parallel processes, the proposed framework effectively reduces computation time without compromising accuracy. The entire system is implemented within the Robot Operating System (ROS) environment \cite{l1}, which facilitates seamless communication between the controller, sensors, and actuators. Specifically, the system integrates data from multiple sensors in real-time, processes the inputs to generate control parameters (acceleration and steering), and actuates the vehicle with minimal latency.

The key contributions of this paper are as follows: \begin{itemize} \item Flexible Speed Control Modes: A method to implement both fixed and variable-speed modes in autonomous driving, enabling the vehicle to adjust its speed dynamically, similar to manual driving behavior. \item Parallelized MPC Computations: A computationally efficient MPC framework that parallelizes next-state prediction, cost function evaluation, and optimization to enhance real-time performance. \item Real-Time Sensor Integration: Real-time fusion of sensor data and integration with ROS to provide accurate control inputs for steering and acceleration, ensuring seamless actuation. \item Reduced Computational Load: The proposed framework significantly reduces the computational load by optimizing resource utilization, enabling smooth and responsive vehicle control in dynamic environments. \end{itemize}

The proposed system is validated through real-time experiments, demonstrating its effectiveness in achieving precise path tracking under varying speed modes. The results highlight the system's ability to adapt to sharp turns, curvatures, and dynamic conditions while maintaining low computational latency and high accuracy.


\section{Related Works}
Nonlinear Model Predictive Control (NMPC) has been extensively explored in recent years for autonomous navigation, path tracking, and collision avoidance due to its ability to handle dynamic constraints and uncertainties. Various studies have demonstrated its versatility in solving complex driving tasks. In~\cite{anil2023trajectory}, Anil \textit{et al.} compared MPC with conventional PID controllers for trajectory tracking, showcasing that MPC achieved superior path-following performance and stability in autonomous driving systems. Similarly, Li \textit{et al.}, in~\cite{li2022design}, proposed an MPC-based navigation framework that incorporates sensor fusion, neural networks, and path exploration, enabling robust performance in uncertain environments.

MPC has also been applied for longitudinal control, addressing smooth speed adjustments while accounting for environmental constraints. In~\cite{mekala2020speed}, Mekala \textit{et al.} used MPC with LiDAR inputs to control vehicle acceleration and deceleration, ensuring a stable response under varying speeds and obstacles. Similarly, Kim \textit{et al.}, in~\cite{kim2023reinforcement}, integrated RL with MPC to optimize real-time highway navigation. Their hybrid approach successfully reduced computational time by over 90\% while maintaining trajectory accuracy, demonstrating the effectiveness of machine learning techniques in augmenting traditional MPC.

Collision avoidance remains a crucial challenge in autonomous systems, and several studies have proposed solutions by integrating MPC with advanced techniques. Yang \textit{et al.}, in~\cite{yang2024collision}, combined an adaptive Artificial Potential Field (APF) method with MPC to generate smooth, collision-free trajectories. By using triangular collision constraints, the computational burden was reduced while maintaining feasibility under dynamic environments. Another notable work, Distributed Cooperative MPC~\cite{mohseni2021distributed} by Mohseni \textit{et al.}, proposed a distributed control approach for multi-vehicle systems. Their framework ensured real-time cooperative lane-switching and intersection maneuvers, highlighting MPC's scalability and efficiency in handling multi-agent control.

To address proactive control, Yoon \textit{et al.} employed Gaussian Process Regression (GPR) in~\cite{yoon2021interaction}. By predicting cut-in vehicle behaviors, their method improved prediction accuracy and enabled the autonomous vehicle to optimize its control strategy, reducing sudden maneuvers. This predictive capability was further extended in~\cite{li2024fast} by Li \textit{et al.}, where computational efficiency was enhanced using dimension reduction strategies, facilitating real-time deployment in dynamic urban environments.

MPC has also been applied for multi-vehicle cooperative systems. Li \textit{et al.}, in~\cite{li2024distributed}, designed a framework that incorporated driver behavior characteristics to ensure safer and more human-like cooperative driving. Similarly, the cooperative approach was expanded in~\cite{mohseni2021distributed}, where distributed strategies enabled vehicles to navigate efficiently in complex traffic situations.

Beyond ground vehicles, MPC applications extend to maritime navigation. Tsolakis \textit{et al.}, in~\cite{tsolakis2024model}, proposed a trajectory optimization method compliant with International Maritime Traffic Rules (COLREGs). Their approach combined rule-based constraints with MPC to ensure collision-free navigation in mixed-traffic maritime environments.

Finally, advancements in multi-agent and dynamic environments were explored in~\cite{li2024fast, kim2023reinforcement}, where RL and MPC integration resulted in computationally efficient and real-time performance. Additionally, APF-based MPC~\cite{yang2024collision} demonstrated high feasibility for dynamic obstacle avoidance scenarios.

The above studies highlight the continuous advancements in MPC for autonomous systems. Whether applied to trajectory tracking, collision avoidance, speed control, or cooperative navigation, MPC remains a critical tool. Recent works emphasize integrating MPC with predictive models, machine learning techniques, and efficient computation strategies to address real-time deployment challenges. This combination of methods enables autonomous vehicles to achieve safety, stability, and efficiency in complex and dynamic environments.

\section{Vehicle Dynamics Model}
\begin{figure}[!h]
  \centering
  \includegraphics[width=0.45\textwidth]{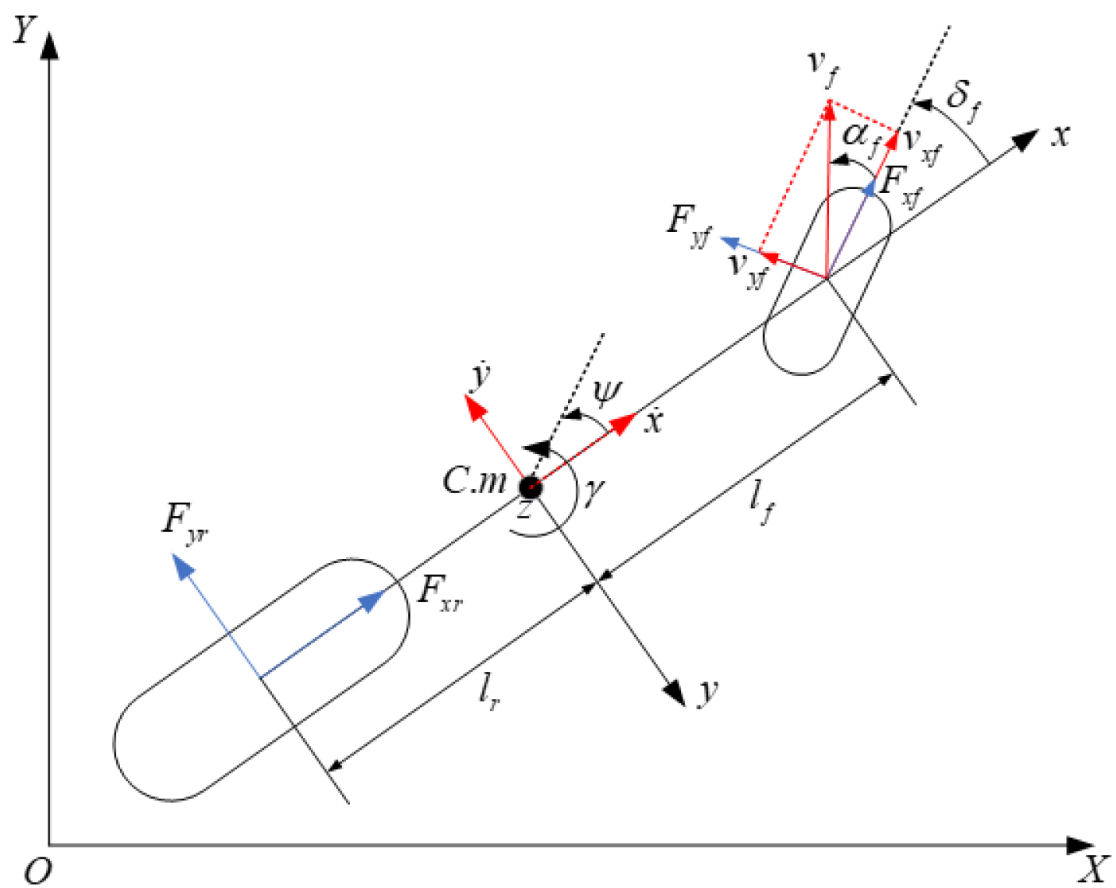}
  \captionsetup{font=small}
  \caption{Geometry of a dynamic bicycle model of a car-like vehicle.}
  \label{bicycle_model}
\end{figure}

The control of an autonomous vehicle requires an accurate representation of its dynamics to predict the states over a finite prediction horizon. In this work, the Linear Parameter Varying (LPV) \cite{lpv1} model is employed due to its capability to approximate nonlinear vehicle dynamics while maintaining computational efficiency. The full vehicle free body diagram is shown in Fig \ref{bicycle_model}

\subsection{State Space Representation}
The vehicle dynamics are expressed in state-space form as:
\[
x(k+1) = A(v) \cdot x(k) + B(v) \cdot u(k),
\]
where:
\begin{itemize}
    \item \(x = [x, y, \psi, v]^T\): State vector,
    \begin{itemize}
        \item \(x, y\): Position in the global frame,
        \item \(\psi\): Heading angle,
        \item \(v\): Longitudinal velocity.
    \end{itemize}
    \item \(u = [a, \delta]^T\): Control input vector,
    \begin{itemize}
        \item \(a\): Acceleration,
        \item \(\delta\): Steering angle.
    \end{itemize}
    \item \(A(v)\) and \(B(v)\): Velocity-dependent system matrices.
\end{itemize}

Here A and B are not constants, rather A is a function of X and B is the function of u. Hence a non-linear model are packed in a linear like model \cite{mpcc12}. All non-linearity are kept in A and B matrix. The control of an autonomous vehicle requires an accurate representation of its dynamics to predict the states over a finite prediction horizon \cite{mpcc11} . In this work, the \textit{Linear Parameter Varying (LPV)} model is employed due to its capability to approximate nonlinear vehicle dynamics while maintaining computational efficiency.

\subsection{Nonlinear Dynamics}
The vehicle's motion is modeled using the kinematic bicycle model,\cite{2_162} expressed as the following equations:

\begin{align}
    \ddot{x} &= a - \mu g - \frac{F_{yf} \sin(\delta)}{m} + \dot{\psi} \dot{y},\label{accel} \\
    \ddot{y} &= \frac{F_{yr}}{m} + \frac{F_{yf} \cos(\delta)}{m} - \dot{\psi} \dot{x}, \\
    \dot{\psi} &= \dot{\psi}, \\
    \ddot{\psi} &= \frac{F_{yf} \cos(\delta) l_f}{I_z} - \frac{F_{yr} l_r}{I_z}, \\
    \dot{X} &= \dot{x} \cos(\psi) - \dot{y} \sin(\psi), \\
    \dot{Y} &= \dot{x} \sin(\psi) + \dot{y} \cos(\psi).
\end{align}

\subsection{Lateral Tire Forces}
The lateral tire forces \( F_{yf} \) and \( F_{yr} \) \cite{2_154} are defined as:
\begin{align}
    F_{yf} &= C_{\alpha f} \left( \delta - \frac{\dot{y}}{\dot{x}} - \frac{\dot{\psi} l_f}{\dot{x}} \right), \\
    F_{yr} &= C_{\alpha r} \left( -\frac{\dot{y}}{\dot{x}} + \frac{\dot{\psi} l_r}{\dot{x}} \right),
\end{align}
where:
\begin{itemize}
    \item \( \delta \): Steering angle,
    \item \( l_f, l_r \): Distances from the center of gravity (CoG) to the front and rear axles,
    \item \( C_{\alpha f}, C_{\alpha r} \): Tire cornering stiffness coefficients,
    \item \( m \): Vehicle mass,
    \item \( I_z \): Yaw moment of inertia,
    \item \( \mu \): Friction coefficient,
    \item \( g \): Gravitational acceleration.
\end{itemize}

\subsection{LPV Representation}
The nonlinear dynamics are linearized around a nominal operating point and expressed in the Linear Parameter Varying (LPV) \cite{lpv2} \cite{lpv3} state-space form:
\begin{align}
    \dot{x} &= A(v) \cdot x + B(v) \cdot u, \\
    y &= C \cdot x,
    \label{state_eq}
\end{align}
where \( x \), \( u \), and \( y \) are the state, input, and output vectors, respectively:
\begin{align}
    x &= \begin{bmatrix} \dot{x} \\ \dot{y} \\ \psi \\ \dot{\psi} \\ X \\Y \end{bmatrix}, \quad
    u = \begin{bmatrix} a \\ \delta \end{bmatrix}, \quad
    y = \begin{bmatrix} \dot{x} \\ \psi \\ X \\Y  \end{bmatrix}.
    \label{state_and_input}
\end{align}

\subsection{System Matrices}
The LPV system matrices \( A(v) \), \( B(v) \), and \( C \) are defined as:

\paragraph{Matrix \( A(v) \)}
\begin{equation}
    \resizebox{\columnwidth}{!}{$
    \begin{bmatrix}
-\frac{\mu g}{\dot{x}} & \frac{C_{\alpha f}\sin(\delta)}{m \dot{x}} & 0 &  \frac{C_{\alpha f} l_f \sin(\delta)}{m \dot{x}} + \dot{y} & 0 & 0 \\
0 & -\frac{C_{\alpha f} + C_{\alpha f}\cos{\delta}}{m \dot{x}} & 0 & -\frac{(C_{\alpha f} l_f \cos{\delta} - C_{\alpha r} l_r)}{m \dot{x}} - \dot{x} & 0 & 0 \\
0 & 0 & 0 & 1 & 0 & 0 \\
0 & -\frac{(C_{\alpha f} l_f \cos{\delta} - C_{\alpha r} l_r)}{I_z \dot{x}} & 0 & -\frac{(C_{\alpha f} l_f^2 \cos{\delta} + C_{\alpha r} l_r^2)}{I_z \dot{x}} & 0 & 0 \\
\cos(\psi) & -\sin(\psi) & 0 & 0 & 0 & 0 \\
\sin(\psi) & \cos(\psi) & 0 & 0 & 0 & 0
\end{bmatrix}.
$}.
\end{equation}

\paragraph{Matrix \( B(v) \)}
\begin{equation}
    B(v) =
    \begin{bmatrix}
         -\frac{C_{\alpha f}\sin(\delta)}{m} & 1 \\
        \frac{C_{\alpha f}\cos(\delta)}{m} & 0 \\
        0 & 0 \\
        \frac{C_{\alpha f} l_f \cos{\delta}}{I_z} & 0\\
        0& 0 \\ 0& 0
    \end{bmatrix}.
\end{equation}

To obtain the output matrix \( C \), the required output should be carefully considered. If only \( \psi \), \( X \), and \( Y \) are selected as the outputs, excessive fluctuations may occur in \( \dot{x} \), which represents the velocity. Therefore, the velocity should also be regulated.
To handle transformations and alignment with the reference trajectory, the velocity components \( \dot{x}_R \) and \( \dot{y}_R \) in the rotated reference frame \( R \) are derived as:
\begin{align}
\dot{x}_R = \dot{x} \cos(\psi_R) - \dot{y} \sin(\psi_R),\\
\dot{y}_R = \dot{x} \sin(\psi_R) + \dot{y} \cos(\psi_R).
\end{align}
These relations ensure that the longitudinal velocity \( \dot{x}_R \) is always tangential to the reference trajectory. \(\dot{y}\) is always perpendicular; hence, it is always \(0\) to track perfectly.

Given the chosen outputs \( \dot{x}, \psi, X, Y \), the matrix \( C \) is:
\begin{equation}
C =
\begin{bmatrix}
1 & 0 & 0 & 0 & 0 & 0 \\
0 & 0 & 1 & 0 & 0 & 0 \\
0 & 0 & 0 & 0 & 1 & 0 \\
0 & 0 & 0 & 0 & 0 & 1
\end{bmatrix}.
\end{equation}

\begin{figure}[!h]
  \centering
  \includegraphics[width=0.48\textwidth]{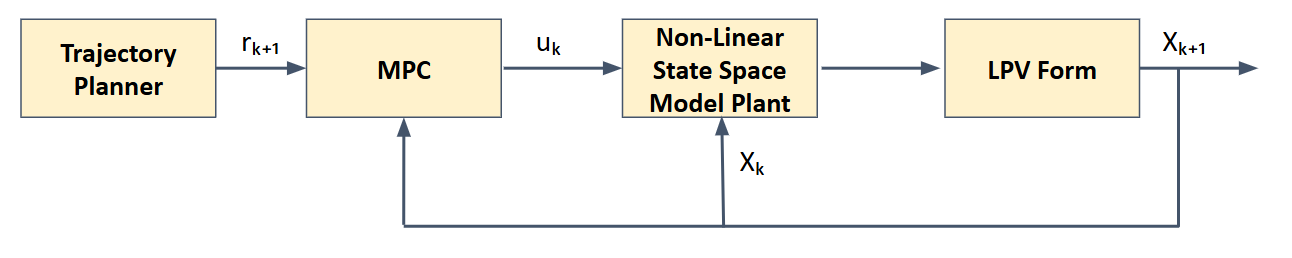}
  \captionsetup{font=small}
  \caption{System Flow Diagram.}
  \label{system_flow}
\end{figure}
The trajectory planner generates reference states \( x_R(t) \) and \( y_R(t) \), which are fed into the MPC. The controller computes the optimal control inputs \( u_k \), ensuring that the states \( x_k \) follow the planned trajectory. The system diagram provided in \ref{system_flow} illustrates this flow, showing how the outputs align with the reference trajectory.

\section{Non-Linear Model Predictive Control Design}
Nonlinear Model Predictive Control (NMPC) \cite{mpcc1} \cite{mpcc2} is a sophisticated control method that determines an optimal set of control inputs over a defined prediction horizon. The aim is to minimize a cost function while adhering to specific system constraints. By utilizing the nonlinear dynamics of the vehicle, NMPC predicts the future states of the system and adjusts control inputs dynamically\cite{mpcc3}. This approach ensures precise trajectory tracking and promotes safe navigation in real-time scenarios.

\subsection{Conversion of State Equations to Discrete Form}
The system dynamics are initially represented in the continuous-time domain as shown in equation \ref{state_eq}. To implement the Model Predictive Control (MPC) framework in discrete time, the equations are converted to discrete form\cite{mpcc4} \cite{mpcc10}. Using a sampling period \( T_s \), the discrete-time equations are expressed as follows:

\begin{align}
\vec{x}_{k+1} = A_d \vec{x}_k + B_d \vec{u}_k,\\
\vec{y}_k = C_d \vec{x}_k + D_d \vec{u}_k,
\label{discrete_state}
\end{align}
where:
\begin{itemize}
    \item \( A_d, B_d, C_d, D_d \) are the discrete-time system matrices.
\end{itemize}







The continuous-time state equations are converted into a discrete-time form suitable for numerical implementation. This process ensures compatibility with the MPC framework,\cite{mpcc5} \cite{mpcc6} which operates over discrete time steps. The discrete system matrices \( A_d \) and \( B_d \) are derived directly from the continuous-time counterparts, providing a seamless transition for predictive control computations.

\subsection{Future State Prediction for MPC}

In Model Predictive Control (MPC), predicting the future states over the horizon period is essential for optimizing control inputs and ensuring the system follows the reference trajectory \cite{mpcc7}. The prediction is based on the discretized state-space equations and iteratively propagates the system dynamics \cite{mpcc5}\cite{mpcc2}.

\subsubsection{State Propagation Equations}
The discrete-time state-space equations are expressed in Eq \ref{discrete_state} .
The future states are predicted by iteratively propagating the state equation:
\begin{itemize}
    \item For \( k = 1 \):
    \[
    \vec{x}_1 = A \vec{x}_0 + B \vec{u}_0
    \]

    \item For \( k = 2 \):
    \[
    \vec{x}_2 = A \vec{x}_1 + B \vec{u}_1 = A^2 \vec{x}_0 + AB \vec{u}_0 + B \vec{u}_1
    \]

    \item For \( k = 3 \):
    \[
    \vec{x}_3 = A^3 \vec{x}_0 + A^2B \vec{u}_0 + AB \vec{u}_1 + B \vec{u}_2
    \]

    \item Generalizing for \( k \):
    \begin{align}
        \vec{x}_k = A^k \vec{x}_0 + \sum_{i=0}^{k-1} A^{k-1-i} B \vec{u}_i
    \end{align}
    
\end{itemize}

\subsubsection{Matrix Form for Horizon Period}

For a prediction horizon \( N \), the future states can be written in a compact form:
\begin{align}
\resizebox{\columnwidth}{!}{$
\begin{bmatrix}
\vec{x}_1 \\
\vec{x}_2 \\
\vdots \\
\vec{x}_N
\end{bmatrix}
=
\begin{bmatrix}
B & 0 & 0 & \dots & 0 \\
AB & B & 0 & \dots & 0 \\
A^2B & AB & B & \dots & 0 \\
\vdots & \vdots & \vdots & \ddots & \vdots \\
A^{N-1}B & A^{N-2}B & A^{N-3}B & \dots & B
\end{bmatrix}
\begin{bmatrix}
\vec{u}_0 \\
\vec{u}_1 \\
\vdots \\
\vec{u}_{N-1}
\end{bmatrix}
+
\begin{bmatrix}
A \\
A^2 \\
\vdots \\
A^N
\end{bmatrix}
\vec{x}_0
$}.
\end{align}

The state propagation over the horizon is visualized as a sequence of predicted states \( \vec{x}_1, \vec{x}_2, \ldots, \vec{x}_N \), which align the system with the reference trajectory. Each predicted state corresponds to a time step in the horizon period. This derivation and compact matrix representation form the basis for computing the optimal control inputs in MPC, ensuring accurate prediction and trajectory tracking over the horizon.

\subsection{Formulating the Cost Function}

The cost function represents the optimization objective for Model Predictive Control (MPC). It is formulated such that it  minimizes the deviation of the system’s output from the reference trajectory while penalizing the control efforts \cite{mpcc8}\cite{mpcc9}. The cost function integrates state errors, input efforts, and the terminal state to ensure smooth, efficient, and accurate control.

\subsubsection{Cost Function Definition}

The cost function \( J \) for a prediction horizon \( N \) is given as:
\begin{align}
J = \frac{1}{2} \vec{e}_{k+N}^T S \vec{e}_{k+N} + \frac{1}{2} \sum_{i=0}^{N-1} \left( \vec{e}_{k+i}^T Q \vec{e}_{k+i} + \vec{u}_{k+i}^T R \vec{u}_{k+i} \right),
\label{cost_function}
\end{align}

where:
\begin{itemize}
    \item \( \vec{e}_{k+i} \): State error at time step \( k+i \),
    \item \( S \): Terminal weight matrix, prioritizing the final state error,
    \item \( Q \): Weight matrix for state tracking errors,
    \item \( R \): Weight matrix for control effort,
    \item \( \vec{u}_{k+i} \): Control input at time step \( k+i \).
\end{itemize}

\subsubsection{State Error Calculation}

The state error \( \vec{e}_{k+i} \) is defined as the difference between the reference state \( \vec{X}_R \) and the measured state \( \vec{X} \):
\begin{align}
\vec{e}_{k+i} = 
\begin{bmatrix}
e_{\dot{x}_{k+i}} \\
e_{\psi_{k+i}} \\
e_{X_{k+i}} \\
e_{Y_{k+i}}
\end{bmatrix}
=
\begin{bmatrix}
\dot{x}_{R_{k+i}} - \dot{x}_{k+i} \\
\psi_{R_{k+i}} - \psi_{k+i} \\
X_{R_{k+i}} - X_{k+i} \\
Y_{R_{k+i}} - Y_{k+i}
\end{bmatrix}.
\end{align}

This error is composed of both dynamic states (e.g., velocity, heading angle) and positional deviations.

\subsubsection{Weight Matrices}

\paragraph{ State Weight Matrix \( Q \)}
The matrix \( Q \) is a diagonal matrix that weights the state error components:
\begin{align}
Q =
\begin{bmatrix}
Q_1 & 0 & 0 & 0 \\
0 & Q_2 & 0 & 0 \\
0 & 0 & Q_3 & 0 \\
0 & 0 & 0 & Q_4
\end{bmatrix}.
\end{align}
Here:
\begin{itemize}
    \item \( Q_1 \): Weight for velocity (\( \dot{x} \)),
    \item \( Q_2 \): Weight for heading angle (\( \psi \)),
    \item \( Q_3, Q_4 \): Weights for \( X \)-position and \( Y \)-position, respectively.
\end{itemize}
Higher weights prioritize reducing specific errors, such as positional deviations, velocity, and heading errors.

\paragraph{ Terminal Weight Matrix \( S \)}
The terminal weight matrix \( S \) penalizes the final state error, ensuring the system converges to the reference trajectory:
\begin{align}
S =
\begin{bmatrix}
S_1 & 0 & 0 & 0 \\
0 & S_2 & 0 & 0 \\
0 & 0 & S_3 & 0 \\
0 & 0 & 0 & S_4
\end{bmatrix}.
\end{align}
Here:
\begin{itemize}
    \item \( S_1 \): Weight for velocity (\( \dot{x} \)) at the terminal state,
    \item \( S_2 \): Weight for heading angle (\( \psi \)) at the terminal state,
    \item \( S_3, S_4 \): Weights for \( X \)-position and \( Y \)-position, respectively, at the terminal state.
\end{itemize}
A larger weight in \( S \) focuses on minimizing the terminal error.

\paragraph{Input Weight Matrix \( R \)}
The control effort is penalized using the matrix \( R \):
\begin{align}
\vec{u}_{k+i}^T R \vec{u}_{k+i} =
\begin{bmatrix}
a_{k+i} & \delta_{k+i}
\end{bmatrix}
\begin{bmatrix}
R_1 & 0 \\
0 & R_2
\end{bmatrix}
\begin{bmatrix}
a_{k+i} \\
\delta_{k+i}
\end{bmatrix}
\end{align}
where:
\begin{itemize}
    \item \( R_1 \): Weight for acceleration (\( a \)),
    \item \( R_2 \): Weight for steering angle (\( \delta \)).
\end{itemize}
These weights ensure that control inputs are kept within reasonable bounds and penalize unnecessary actuation.

Expanding the summation term in the cost function ensures that errors and inputs are minimized across the entire prediction horizon, with the terminal state given higher priority through \( S \). This comprehensive cost function ensures that the system effectively tracks the reference trajectory while maintaining smooth control actions, ultimately achieving stability and performance over the prediction horizon.

\subsection{Finalizing Constraints}

Constraints ensure that the system operates within safe and physically feasible boundaries. These constraints are applied to control inputs, states, and state derivatives, maintaining stability and compliance with vehicle dynamics.

\subsubsection{Fixed Constraints}
The following fixed constraints are applied to the system:
\begin{enumerate}
    \item Steering rate (\( \Delta \delta \)):
    \begin{align}
        -\frac{\pi}{300} \leq \Delta \delta \leq \frac{\pi}{300},
    \end{align}

    \item Acceleration rate (\( \Delta a \)):
    \begin{align}
    -0.1 \leq \Delta a \leq 0.1 \, \text{m/s}^2.
    \end{align}
    \item Steering angle (\( \delta \)):
    \begin{align}
    -\frac{\pi}{6} \leq \delta \leq \frac{\pi}{6}.
    \end{align}
    
\end{enumerate}

\subsubsection{Constraint on Acceleration}
Dynamic constraints are introduced to handle acceleration (\( a \)) by using the total net acceleration (\( \ddot{x} \)) derived from the system dynamics given in Eq \ref{accel},
 the bounds for \( \ddot{x} \) are:
 \begin{align}
-4 \, \text{m/s}^2 \leq \ddot{x} \leq 2 \, \text{m/s}^2.
 \end{align}
Substituting \( \ddot{x} \) into the equation \ref{accel} gives:
\begin{align}
-4 + \frac{F_{yf} \sin(\delta)}{m} + \mu g - \dot{\psi} \dot{y} \leq a \leq 1 + \frac{F_{yf} \sin(\delta)}{m} + \mu g - \dot{\psi} \dot{y}.
\end{align}

This forms the changing constraints on acceleration (\( a \)) based on dynamic conditions.

\subsubsection{Constraints on Velocity and Lateral Velocity}
The longitudinal velocity (\( \dot{x} \)) and lateral velocity (\( \dot{y} \)) are constrained as follows:
\begin{enumerate}
    \item For longitudinal velocity (\( \dot{x} \)):
    \begin{align}
    1 \, \text{m/s} \leq \dot{x} \leq 30 \, \text{m/s}.
    \end{align}

    \item For lateral velocity a changing constraint is applied to \( \dot{y} \), ensuring that lateral movement is proportional to longitudinal velocity (\( \dot{x} \)):
    \begin{align}
-0.17 \dot{x} \leq \dot{y} \leq 0.17 \dot{x}.
    \end{align}
This is further limited by the fixed constraint of \( \dot{y} \).
\end{enumerate}
Combining all the fixed and changing constraints, the system constraints are tabulated in Table \ref{constraints}

\begin{table}[h!]
\centering
\caption{Summary of Constraints for MPC}
\label{constraints}
\renewcommand{\arraystretch}{1.5} 
\resizebox{\columnwidth}{!}{
\begin{tabular}{|c|c|}
\hline
\textbf{Constraint Type}         & \textbf{Range/Limit}                                                              \\ \hline
Steering rate (\( \Delta \delta \)) & \( -\frac{\pi}{300} \leq \Delta \delta \leq \frac{\pi}{300} \)             \\ \hline
Acceleration rate (\( \Delta a \)) & \( -0.1 \leq \Delta a \leq 0.1 \, \text{m/s}^2 \)                                  \\ \hline
Steering angle (\( \delta \))      & \( -\frac{\pi}{6} \leq \delta \leq \frac{\pi}{6} \)                        \\ \hline
Acceleration (\( a \))             & \( -4 + \frac{F_y^f \sin(\delta)}{m} + \mu g - \dot{\psi} \dot{y} \leq a \leq 1 + \frac{F_y^f \sin(\delta)}{m} + \mu g - \dot{\psi} \dot{y} \) \\ \hline
Longitudinal velocity (\( \dot{x} \)) & \( 1 \leq \dot{x} \leq 30 \, \text{m/s} \)                                       \\ \hline
Lateral velocity (\( \dot{y} \))    & \(  -0.17 \dot{x} \leq \dot{y} \leq 0.17 \dot{x} \) \\ \hline
\end{tabular}}
\end{table}

\subsection{Solving the Cost Function with CasADi}

Solving the cost function involves optimizing the control inputs to minimize the state error while adhering to system dynamics and constraints. CasADi,\cite{cas1} a powerful symbolic framework for numerical optimization, is utilized for this purpose. It efficiently formulates and solves the constrained optimization problem inherent in MPC.

CasADi is used to symbolically define and solve the optimization problem, and the process involves the following steps\cite{cas1}:

\paragraph{Defining the State and Input Variables}
The state vector \( \vec{x} \) and control input \( \vec{u} \) are defined symbolically in equation \ref{state_and_input}.

\paragraph{Prediction Model}
The state dynamics are iteratively computed over the horizon \( N \) using the discrete-time system equations as derived in equation \ref{discrete_state}. CasADi \cite{cas2} allows for compact symbolic computation of predicted states over the horizon:
\[
\vec{x}_{k+1} = f(\vec{x}_k, \vec{u}_k), \quad \forall k \in \{0, 1, \dots, N-1\}.
\]

\paragraph{Cost Function Construction}
The cost function is built iteratively for each prediction step \( i \) as shown in equation \ref{cost_function}.

\paragraph{Defining Constraints}
The constraints are symbolically defined as given in table \ref{constraints}.

\paragraph{Optimization Problem}
The entire problem is formulated as:
\[
\text{Minimize } J(\vec{u}_0, \dots, \vec{u}_{N-1}) \quad \text{subject to:}
\]
\begin{itemize}
    \item System dynamics:
    \[
    \vec{x}_{k+1} = A_d \vec{x}_k + B_d \vec{u}_k.
    \]
    \item Constraints.
\end{itemize}

CasADi provides interfaces for numerical solvers such as IPOPT to solve the optimization problem. The solver iteratively computes the control inputs \( \vec{u}_k \) that minimize \( J \) while satisfying all constraints and gives optimal output \(y_k\).


The result of the optimization is the sequence of control inputs:
\[
\vec{u}_k =
\begin{bmatrix}
a_k & \delta_k
\end{bmatrix}.
\]
At each time step, only the first control input \( \vec{u}_0 \) is applied, and the optimization problem is re-solved in a receding horizon manner.

CasADi offers a robust framework for symbolic modeling and efficient numerical optimization, making it ideal for implementing MPC. Its ability to handle complex constraints and nonlinear dynamics ensures optimal performance and real-time feasibility, providing precise control inputs for trajectory tracking and system stability.


\section{Implementation Framework}

The development of the Model Predictive Control (MPC) framework involved two stages: simulation and real-time implementation. A Python-based simulation platform with real-time animation was designed to validate the algorithm before deployment. The simulation incorporated exact vehicle parameters, where a virtual vehicle model was developed. The same state estimation and MPC controller were applied within this platform, allowing results to be observed in a controlled environment. Following successful simulation testing, the algorithm was implemented on a real-time system.

For real-time testing, a Novatel GPS sensor was used to acquire position and IMU information, ensuring precise localization. The platform utilized an M2 category drive-by-wire vehicle, which provided control over steering, braking, and acceleration. Additionally, all computations, including state estimation and MPC optimization, were executed on a Nvidia Jetson AGX Orin Developer Kit, equipped with a 64GB memory and a 12-core ARM Cortex CPU. This platform was used for both the simulation and real-time implementations, enabling consistency and seamless transitions between testing environments.

\begin{figure}[!h]
  \centering
  \includegraphics[width=0.45\textwidth]{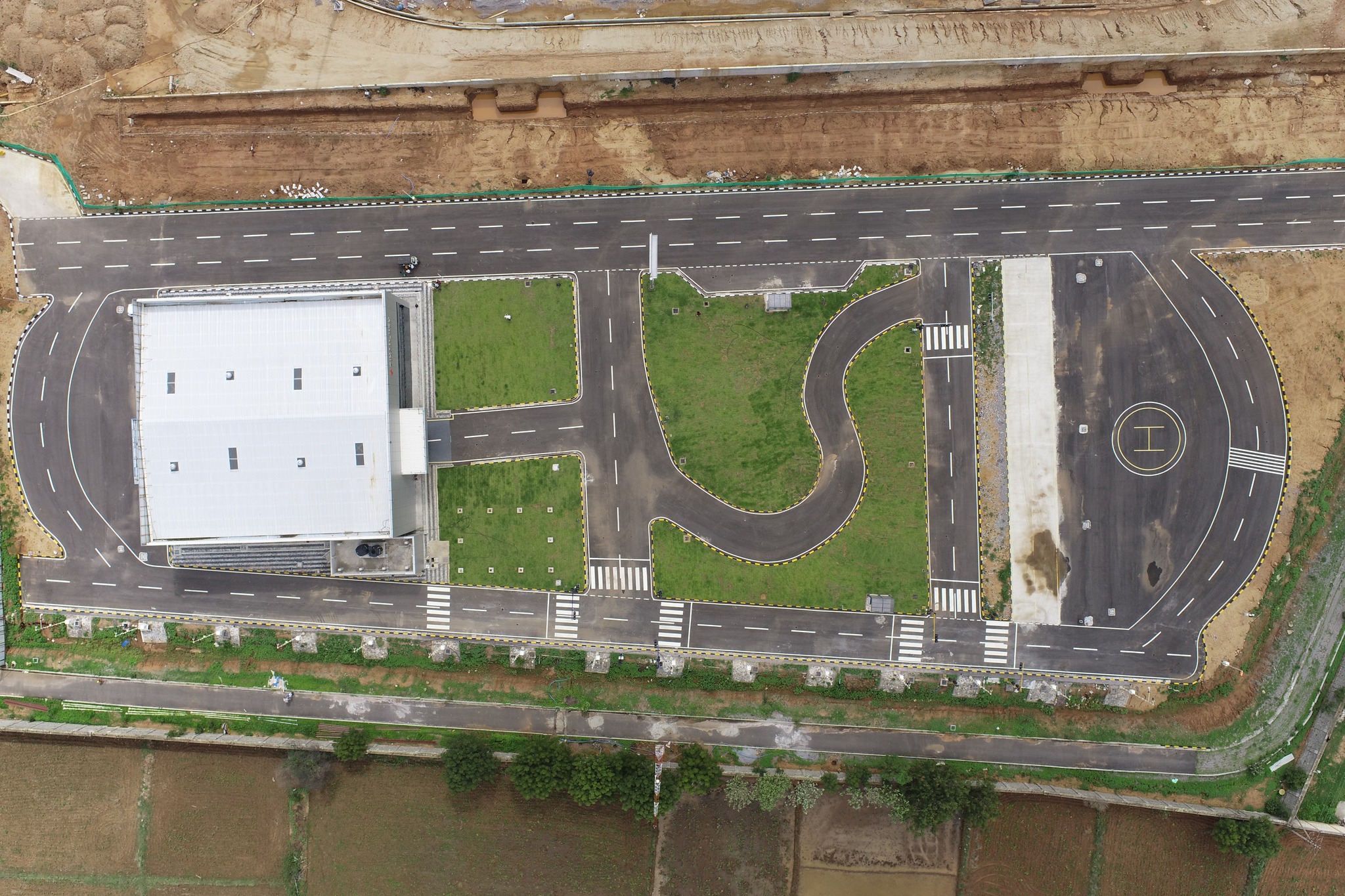}
  \captionsetup{font=small}
  \caption{TiHAN Test-track}
  \label{tihan_testbed}
\end{figure}

All the testings are performed in the campus of Indian Institute of Technology (IIT) Hyderabad and in Technology Innovation Hub on Autonomous Navigation (TiHAN) testbed. The TiHAN (Technology Innovation Hub on Autonomous Navigation and Data Acquisition Systems) \cite{testbed} \cite{tihan} at IIT Hyderabad is India’s first multidisciplinary testbed dedicated to autonomous navigation. Fig \ref{tihan_testbed} shows the test-track of TiHAN. TiHAN provides cutting-edge facilities for the testing and development of autonomous vehicles, drones, and other navigation systems. Its state-of-the-art infrastructure includes specialized test tracks, multimodal sensors (radar, LiDAR, cameras, GPS), and capabilities for V2X communication and real-time data acquisition. TiHAN focuses on advancing technologies such as sensor fusion, localization, SLAM, AI-based navigation, and control algorithms, offering a robust platform to innovate and test autonomous systems in real-world scenarios.

\begin{figure*}[!h]
  \centering
  \includegraphics[width=1.0\textwidth, height=0.32\textwidth]{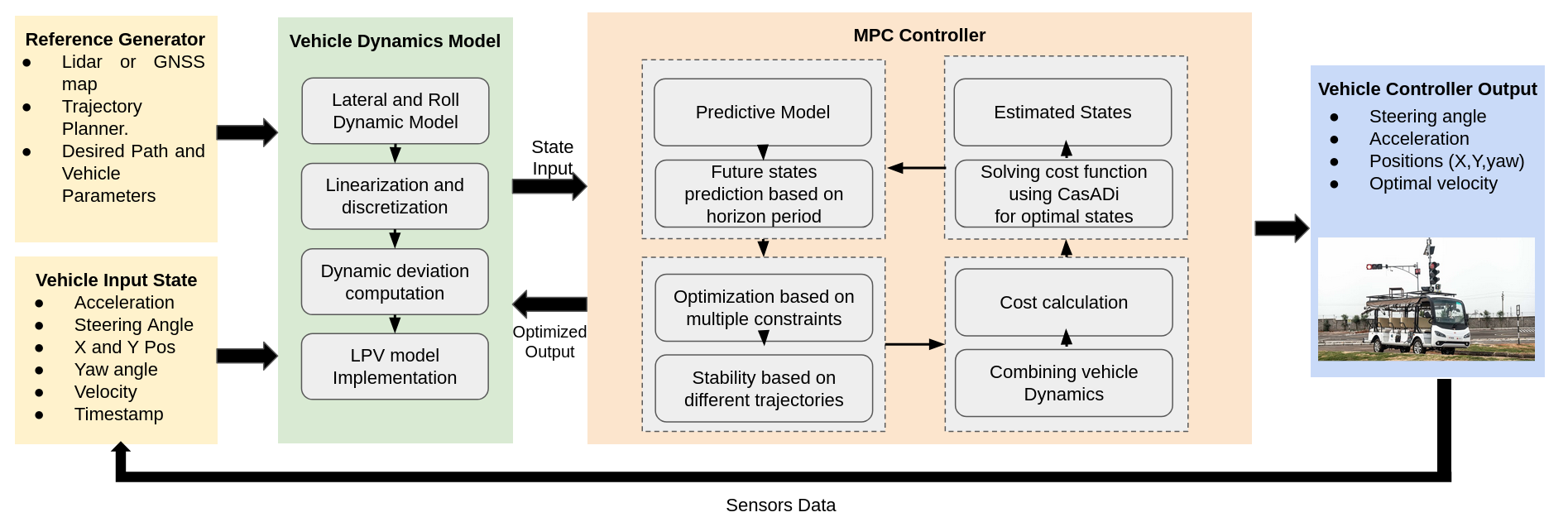}
  \captionsetup{font=small}
  \caption{Overall MPC Flow Architecture}
  \label{MPC_arch}
\end{figure*}

The implementation of the MPC framework is realized through a modular architecture that integrates state estimation, optimization, and actuation. The process is orchestrated using the Robot Operating System (ROS) for real-time data handling and parallel computation. The block diagram in Fig \ref{MPC_arch} illustrates the overall framework and the interconnections between its key components.

\subsection{Overview of the Framework}
The implementation framework begins with a Reference Generator, which provides the desired path, yaw angle, and vehicle parameters. The vehicle's state is estimated in real-time using data from various onboard sensors, including GPS, IMU, and velocity sensors. This state data is processed and refined to ensure accuracy, forming the input for the Vehicle Dynamics Model and the MPC Controller.

The Vehicle Dynamics Model incorporates lateral and roll dynamics, linearization, and an LPV (Linear Parameter Varying) representation of the system. This model predicts the vehicle’s behavior and provides essential inputs for the MPC optimization process. The MPC Controller uses this information to compute future states and optimal control inputs, ensuring trajectory tracking and stability.

The framework is designed to operate in real-time, with parallel execution of state estimation and MPC optimization in separate ROS nodes. This parallelism ensures minimal latency and efficient computation, critical for autonomous driving applications.

\subsection{ROS Architecture}
The entire implementation leverages the ROS framework for modularity, scalability, and efficient communication between components. Key features of the ROS-based architecture include:

\begin{enumerate}
    \item \textbf{Modular Nodes:}
    \begin{itemize}
        \item State estimation and MPC optimization are implemented as independent ROS nodes, enabling concurrent processing.
        \item Each node is responsible for a specific task, such as sensor data processing, vehicle state estimation, or control optimization.
    \end{itemize}
    
    \item \textbf{Data Flow:}
    \begin{itemize}
        \item Sensor data is published as ROS topics, which are subscribed to by the state estimation node. This node processes the data to produce refined vehicle states, including position, orientation, and velocity.
        \item The estimated state is fed into the MPC optimization node, which solves the cost function and generates optimal control inputs for acceleration and steering.
    \end{itemize}
    
    \item \textbf{Communication:}
    \begin{itemize}
        \item ROS topics facilitate seamless communication between nodes, ensuring real-time transfer of state data and optimized control inputs.
        \item The output from the MPC node, consisting of the optimized control parameters, is fed to the actuation system to control the vehicle.
    \end{itemize}
\end{enumerate}

\subsection{Tuning Weight Matrices}
In this framework, the weight matrix \( Q \) is responsible for penalizing deviations in state variables, such as position, orientation, and velocity, while \( S \) penalizes terminal state errors to ensure convergence to the reference trajectory. The tuning process begins in the simulation environment, where the effects of different weight values are systematically evaluated. The vehicle's ability to follow the reference trajectory with minimal overshoot, oscillation, or steady-state error is used as the primary metric for tuning.

The tuning of weight matrices \( Q \) and \( S \) was guided by the curvature analysis of the trajectory. The curvature (\( \kappa \)) at a point on a 2D trajectory is a measure of how sharply the curve bends \cite{curv1}\cite{curv2} and is given by:

\begin{equation}
    \kappa = \frac{|x'y'' - y'x''|}{(x'^2 + y'^2)^{3/2}},
\end{equation}

where \( x' \) and \( y' \) are the first derivatives of \( x \) and \( y \) with respect to the parameter \( s \) (distance along the trajectory), and \( x'' \) and \( y'' \) are the second derivatives of \( x \) and \( y \). The parameter \( s \), which represents the cumulative arc length along the trajectory, is calculated as \cite{curv4}:

\begin{equation}
    s_i = \sum_{j=1}^{i} \sqrt{(x_j - x_{j-1})^2 + (y_j - y_{j-1})^2}.
\end{equation}

- \( x \) and \( y \): Typically in meters (\( m \)).
- \( \kappa \): Reciprocal of distance, typically in \( m^{-1} \).

- **High curvature (\( \kappa \))**: Indicates a sharp turn.
- **Low curvature (\( \kappa \))**: Indicates a gentle or straight path \cite{curv3}.

The weight matrix \( Q \) was tuned based on the sharpest curvature (\( \kappa_{\text{max}} \)) observed in a path. Higher weights were assigned to penalize deviations in critical states, such as lateral position or heading angle, during sharp turns. Similarly, the terminal weight matrix \( S \) was adjusted to ensure trajectory convergence near regions of high curvature. The tuning process involved:

1. Simulation-Based Tuning:
    - Curvature data (\( \kappa \)) was used to classify the path into regions of varying sharpness.
    - Weight parameters were iteratively adjusted based on the vehicle's performance in following the trajectory.

2. Generalized Framework:
    - A generalized framework was developed to autotune the weight matrices \( Q \) and \( S \) based on the sharpest curvature of a given path.
    - For predefined curvature thresholds, corresponding weight matrices were applied, ensuring robust performance across a range of trajectories.

By integrating this curvature-based tuning mechanism, the MPC controller demonstrated improved trajectory tracking, especially in sharp turns, while maintaining stability and control effort within acceptable limits.

\subsection{Actuation and Feedback}
The optimized outputs from the MPC controller are passed to the actuation system, where they are converted into physical commands for steering and acceleration. A feedback loop ensures that the updated vehicle state is continuously monitored and refined, enabling closed-loop control.

This implementation framework demonstrates the efficacy of a modular, ROS-based approach for deploying MPC in autonomous vehicles. By leveraging real-time data handling, parallel processing, and advanced optimization techniques, the system achieves robust trajectory tracking and control.



\begin{figure*}[!h]
    \centering
    \includegraphics[width=1.0\textwidth]{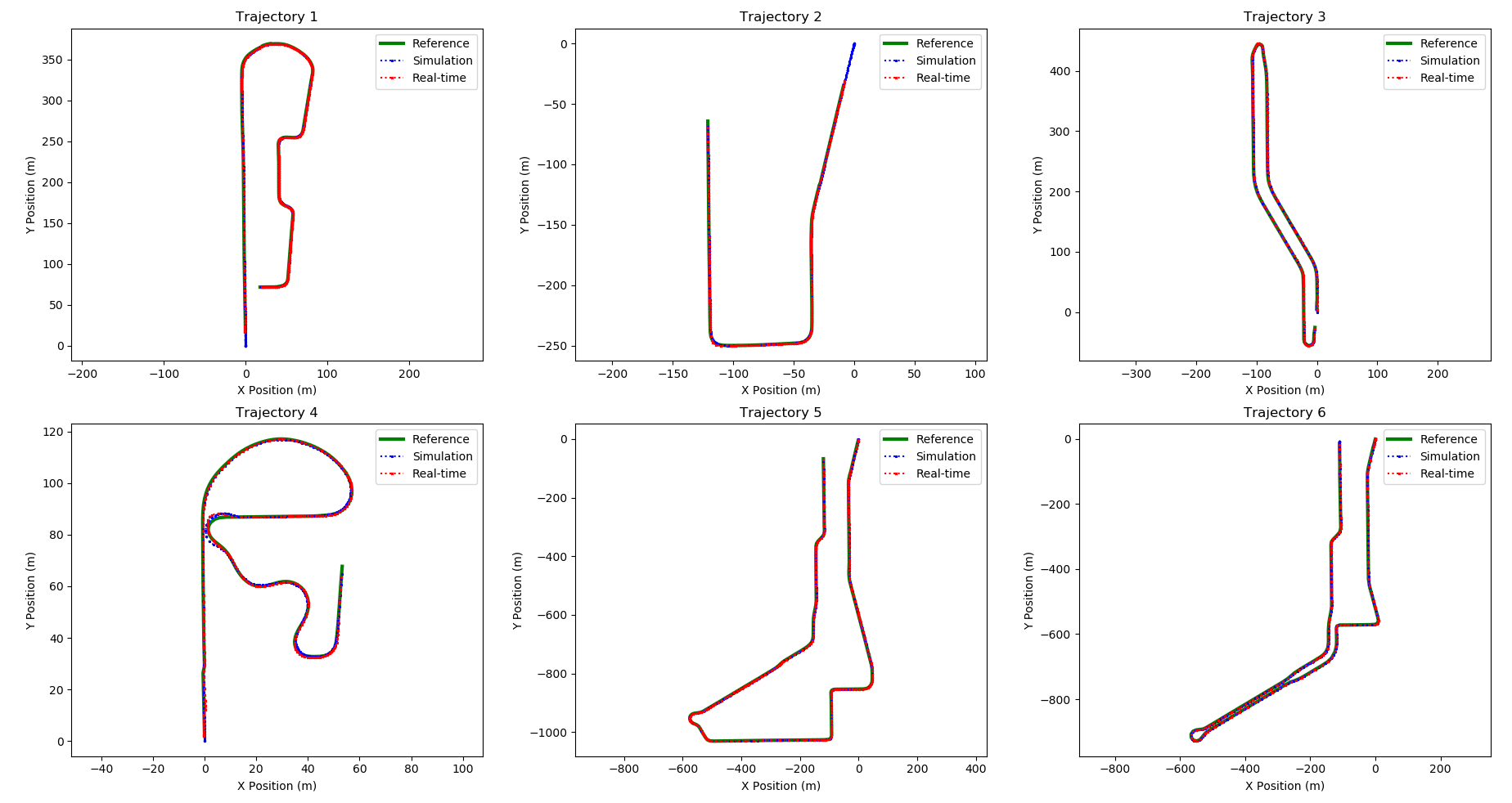}
    \captionsetup{font=small}
    \caption{Comparison of reference, simulation, and real-time trajectories for six test cases.}
    \label{trajectory_plots}
\end{figure*}

\section{Results and Discussion}

The performance of the Model Predictive Control (MPC) algorithm is evaluated in both simulation and real-time environments across six different trajectories. The results are presented in Fig. \ref{trajectory_plots}, which compares the deviation from the reference path in simulation and real-time execution. These trajectories cover a range of driving scenarios, including quiet, moderate, high-speed, and aggressive driving conditions. The table presented in Table \ref{trajectory_info} summarizes the details of the trajectories used for optimization and testing purposes.
\begin{table}[!h]
    \centering
    \captionsetup{font=small}
    \caption{Trajectories used in optimization and experimental tests.}
    \label{trajectory_info}
    \renewcommand{\arraystretch}{1.3} 
    \setlength{\tabcolsep}{0.04cm} 
    \small 
    \begin{tabular}{|c|c|c|c|c|c|c|}
        \hline
        \textbf{Trajectory} & \textbf{T1} & \textbf{T2} & \textbf{T3} & \textbf{T4} & \textbf{T5} & \textbf{T6} \\
        \hline
        Maximum speed (km/h) & 50.36 & 28.05 & 31.01 & 23.01 & 38.59 & 28.12 \\
        \hline
        \begin{tabular}[c]{@{}c@{}}Max. long. \\ Acceleration (m/s\(^2\))\end{tabular} & 1.32 & 2.01 & 1.51 & 2.0 & 1.85 & 1.86 \\
        \hline
        \begin{tabular}[c]{@{}c@{}}Max. lat. \\ Acceleration (m/s\(^2\))\end{tabular} & 1.1 & 1.5 & 1.89 & 1.2 & 1.56 & 2.0 \\
        \hline
        Length (m) & 783.51 & 513.17 & 1033.47 & 500.0 & 2119.6 & 1959.3 \\
        \hline
        Max Curvature (m\(^{-1}\)) & 1.81 & 2.79 & 3.98 & 4.0 & 4.0 & 2.0 \\
        \hline
        Total Curvature (m\(^{-1}\)) & 42.57 & 24.49 & 44.35 & 44.0 & 45.0 & 20.0 \\
        \hline
    \end{tabular}
\end{table}

The plots in Fig \ref{trajectory_plots} demonstrate the capability of the MPC framework to follow the reference path with minimal deviation in both simulation and real-time scenarios. The slight variations observed between the real-time and simulation trajectories are attributed to sensor noise, actuator dynamics, and external environmental factors.

\begin{figure}[!h]
  \centering
  \includegraphics[width=0.5\textwidth]{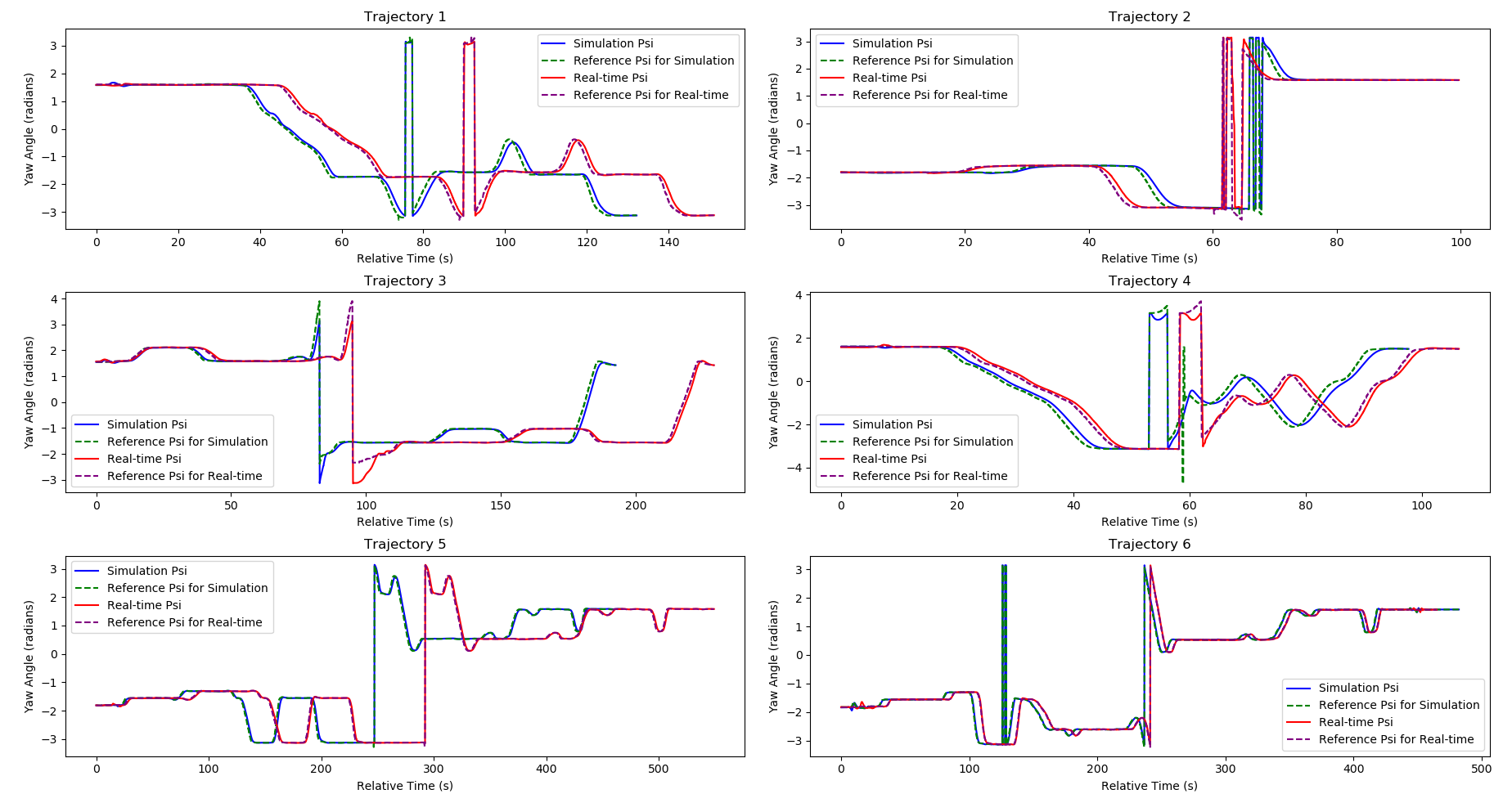}
  \captionsetup{font=small}
  \caption{Yaw angle tracking performance for six trajectories in simulation and real-time environments. The plot compares the yaw angle (\(\psi\)) and its reference for both simulation and real-time execution.}
  \label{yaw_tracking}
\end{figure}

Figure \ref{yaw_tracking} illustrates the yaw angle tracking performance for six trajectories. The yaw angle (\( \psi \)) is tracked against its reference values for both simulation and real-time execution. The following observations can be made:



\begin{table*}[!h]
\centering
\renewcommand{\arraystretch}{1.2}
\caption{Performance Metrics for Trajectories in Simulation and Real-Time}
\label{table:performance_metrics}
\resizebox{\textwidth}{!}{%
\begin{tabular}{|c|cccccc|cccccc|}
\hline
\multirow{2}{*}{\textbf{Trajectory}} & \multicolumn{6}{c|}{\textbf{Simulation}}                                                                                                                                         & \multicolumn{6}{c|}{\textbf{Real-Time}}                                                                                                                                          \\ \cline{2-13} 
                                     & \textbf{Max CTE (m)} & \textbf{Mean CTE (m)} & \textbf{MLE (m)} & \textbf{ALE (m)} & \textbf{MOE (rad)} & \textbf{AOE (rad)} & \textbf{Max CTE (m)} & \textbf{Mean CTE (m)} & \textbf{MLE (m)} & \textbf{ALE (m)} & \textbf{MOE (rad)} & \textbf{AOE (rad)} \\ \hline
\textbf{T1}                          & 1.36                 & 0.25                  & 1.35             & 0.19             & 0.45         & 0.08         & 2.86                 & 0.26                  & 2.71             & 0.20             & 0.57         & 0.08         \\ \hline
\textbf{T2}                          & 0.76                 & 0.17                  & 0.73             & 0.14             & 0.63         & 0.04         & 1.97                 & 0.19                  & 1.67             & 0.16             & 0.77         & 0.03         \\ \hline
\textbf{T3}                          & 0.99                 & 0.19                  & 0.97             & 0.15             & 0.57         & 0.05         & 2.23                 & 0.21                  & 2.13             & 0.17             & 0.65         & 0.05         \\ \hline
\textbf{T4}                          & 1.73                 & 0.30                  & 1.61             & 0.20             & 0.54         & 0.18         & 2.09                 & 0.23                  & 1.92             & 0.17             & 0.63         & 0.13         \\ \hline
\textbf{T5}                          & 0.82                 & 0.19                  & 0.82             & 0.13             & 0.37         & 0.03         & 2.66                 & 0.18                  & 2.58             & 0.12             & 0.42         & 0.03         \\ \hline
\textbf{T6}                          & 0.77                 & 0.16                  & 0.72             & 0.11             & 0.29         & 0.02         & 0.90                 & 0.19                  & 0.79             & 0.13             & 0.18         & 0.01         \\ \hline
\end{tabular}%
}
\label{erroe_tab}
\end{table*}

\begin{enumerate}
    \item \textbf{Accuracy:} Across all trajectories, the yaw angle closely follows the reference values, demonstrating the effectiveness of the Model Predictive Controller (MPC).
    \item \textbf{Deviations:} Minor deviations are observed in Trajectories 2 and 5 due to higher curvature sections, which present challenges for real-time actuation.
    \item \textbf{Real-time vs. Simulation:} The yaw angle tracking in real-time aligns well with simulation results, showcasing the robustness of the MPC implementation across diverse scenarios.
\end{enumerate}
The yaw angle tracking performance validates the ability of the MPC to ensure stability and accuracy under varying driving conditions. The close alignment between real-time and simulation results indicates that the developed control framework effectively transitions from simulation to real-world applications. Higher curvatures in the trajectory affect the real-time tracking slightly more than the simulation due to actuation delays, which can be addressed through fine-tuning of the weight matrices or actuation system.

\subsection{Performance Criteria}
Controller performance is quantitatively assessed using the following metrics\cite{rev2} \cite{IET}:

\begin{itemize}
    \item Maximum lateral error (MLE):
    \begin{equation}
    e_{d,\text{max}} = \max_{t \in [0, T]} |e_d(t)|
    \end{equation}
    
    \item Maximum orientation error (MOE):
    \begin{equation}
    e_{\theta,\text{max}} = \max_{t \in [0, T]} |e_\theta(t)|
    \end{equation}

    \item Average lateral error (ALE):
    \begin{equation}
    e_{d,\text{rms}} = \sqrt{\frac{1}{T} \int_0^T e_d(t)^2 dt}
    \end{equation}

    \item Average orientation error (AOE):
    \begin{equation}
    e_{\theta,\text{rms}} = \sqrt{\frac{1}{T} \int_0^T e_\theta(t)^2 dt}
    \end{equation}
    
    \item Cross-Track Error (CTE), evaluated for different path representations:
    \begin{itemize}
        \item For a straight reference path $ax + by + c = 0$:
        \begin{equation}
        \text{CTE} = \frac{|ax_v + by_v + c|}{\sqrt{a^2 + b^2}}
        \end{equation}
        
        \item For a path given as a function $y = f(x)$:
        \begin{equation}
        \text{CTE} = y_v - f(x_v)
        \end{equation}
        
        \item For parametric or curved paths:
        \begin{equation}
        \text{CTE} = (x_v - x_r^\ast) \sin(\theta_r) - (y_v - y_r^\ast) \cos(\theta_r)
        \end{equation}
    \end{itemize}
\end{itemize}

The table \ref{erroe_tab} summarizes the performance of the MPC framework across six trajectories in both simulation and real-time environments. Key observations include:

\begin{itemize}
    \item \textbf{Cross-Track Error (CTE):} 
    - In simulation, the maximum CTE (Max CTE) values are significantly lower than in real-time due to the absence of external disturbances and actuation delays. 
    - Real-time Max CTE is highest for Trajectory T1 (2.86 m) due to sharp turns and aggressive maneuvers.

    \item \textbf{Mean CTE:} 
    - Mean CTE remains consistently low across all trajectories in both simulation and real-time, indicating effective path tracking by the MPC controller.

    \item \textbf{Lateral Errors (MLE and ALE):} 
    - Lateral errors (MLE and ALE) show slight deviations in real-time compared to simulation due to vehicle dynamics and latency in actuation.

    \item \textbf{Orientation Errors (MOE and AOE):} 
    - Orientation errors remain minimal in both environments, with AOE in simulation consistently lower than real-time, validating the accuracy of the yaw angle tracking.

    \item \textbf{Real-Time Challenges:}
    - Real-time execution faces challenges such as sensor noise, actuation delays, and environmental disturbances, leading to higher errors compared to simulation.
    - The results still demonstrate robust performance, with acceptable deviations from the reference trajectory.

\end{itemize}
These metrics validate the efficacy of the MPC framework in achieving accurate trajectory tracking in diverse scenarios while highlighting areas for further improvement, particularly in real-time implementations.

\begin{figure}[!h]
  \centering
  \includegraphics[width=0.5\textwidth]{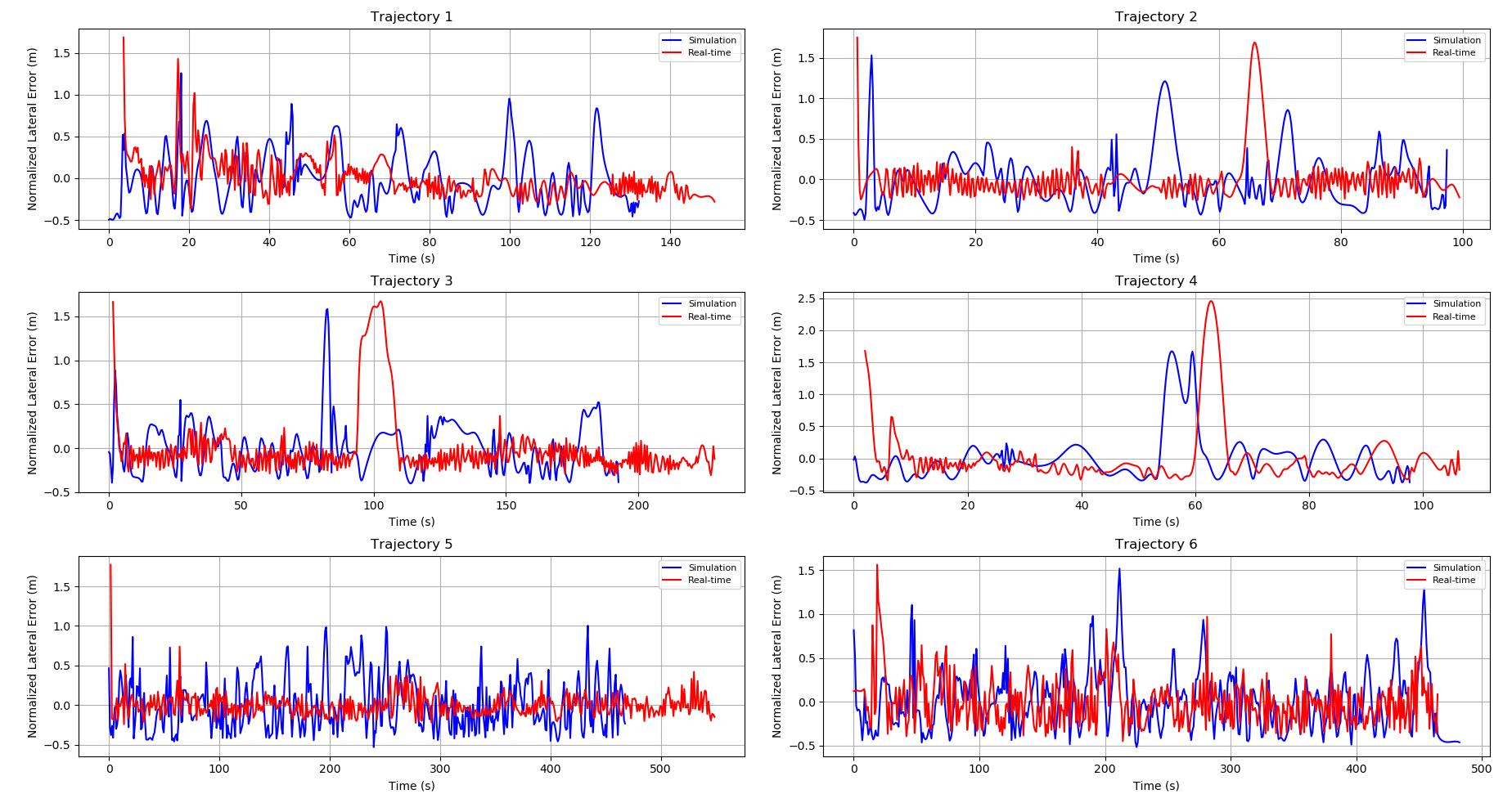}
  \captionsetup{font=small}
  \caption{Lateral error plots for all trajectories comparing simulation and real-time results}
  \label{lateral_error}
\end{figure}

The lateral error for all six trajectories in both simulation and real-time scenarios is depicted in Figure \ref{lateral_error}. The graph showcases the normalized lateral error over time for each trajectory, comparing the simulated and real-time performance of the Model Predictive Controller (MPC).

\begin{figure}[!h]
  \centering
  \includegraphics[width=0.5\textwidth]{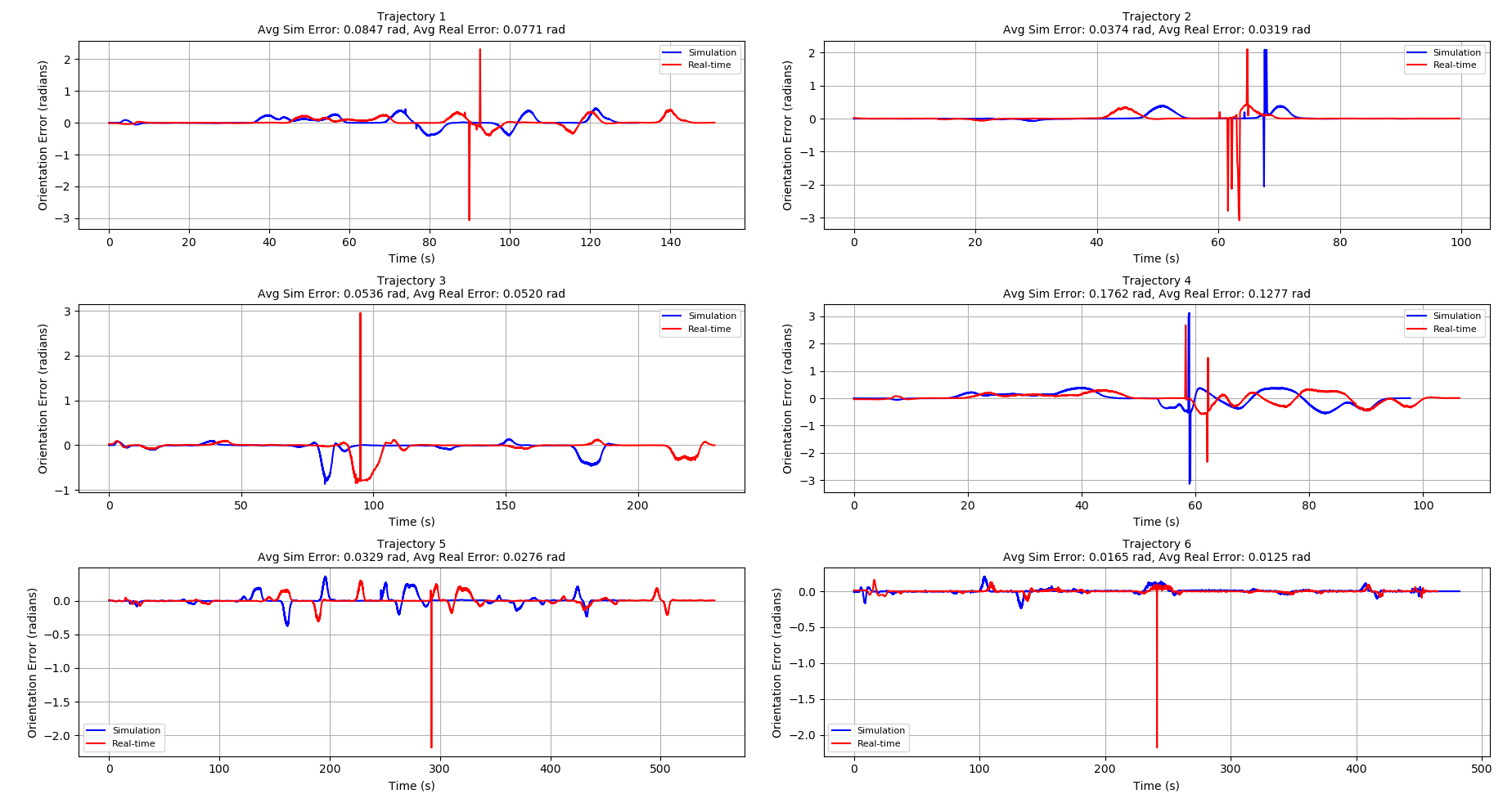}
  \captionsetup{font=small}
  \caption{Orientation error plots for all trajectories comparing simulation and real-time results}
  \label{orientation_error}
\end{figure}

In The lateral and orientation error in Fig \ref{lateral_error} and Fig \ref{orientation_error} is well-contained in both simulation and real-time scenarios, with occasional spikes observed in real-time due to environmental disturbances or sensor noise.

Trajectory 5 and Trajectory 6 exhibit more significant lateral error deviations in real-time, especially during prolonged segments with high curvature. This is consistent with the system's response to aggressive maneuvers or high-speed operation.

The simulation performance aligns closely with the reference trajectory, indicating the effectiveness of the MPC in ideal conditions. In real-time scenarios, the controller maintains robust performance despite larger deviations, highlighting the system's ability to adapt to dynamic and uncertain environments. The differences between simulation and real-time results underscore the impact of sensor noise, vehicle dynamics, and external disturbances on the controller's accuracy.

\section{Conclusion}
This research has successfully demonstrated the application of a Model Predictive Control (MPC) framework for autonomous vehicle trajectory tracking under diverse driving conditions. The proposed architecture integrates state estimation, optimization, and actuation modules, with parallelized computation using the Robot Operating System (ROS), enabling robust real-time performance. Both simulation and real-time experiments validated the controller’s capability to achieve precise trajectory tracking with minimal cross-track error (CTE) and orientation error, even in scenarios involving sharp turns and aggressive maneuvers. The integration of curvature-based tuning of weight matrices further enhanced the controller’s ability to handle challenging trajectories, showcasing adaptability and robustness. The results indicate that the developed MPC framework is effective in aligning simulation and real-time performance while maintaining reliable trajectory tracking under varied driving scenarios, including high-speed and sharp-turn conditions.

While the developed MPC framework exhibits strong performance in diverse scenarios, several opportunities exist to extend its applicability and robustness. Future efforts could focus on dynamically adapting the weight matrices in real-time based on environmental conditions and sensor feedback to improve the system's adaptability to unknown scenarios. The integration of the framework in multi-vehicle cooperative navigation systems for efficient traffic management and collision avoidance also presents an interesting direction for exploration. Incorporating additional sensors, such as LiDAR and radar, could further enhance state estimation accuracy and improve the system's resilience to environmental disturbances. These advancements could further elevate the proposed MPC framework into a comprehensive solution for safe and efficient autonomous vehicle navigation in complex, dynamic scenarios.

\section{Acknowledgment}
This work is supported by DST National Mission Interdisciplinary Cyber-Physical Systems (NM-ICPS), Technology Innovation Hub on Autonomous Navigation and Data Acquisition Systems: TiHAN Foundation at Indian Institute of Technology (IIT) Hyderabad.


\section{Biography Section}

\vspace{11pt}

\vspace{-33pt}
\begin{IEEEbiography}
[{\includegraphics[width=1in,height=1.25in,clip,keepaspectratio]{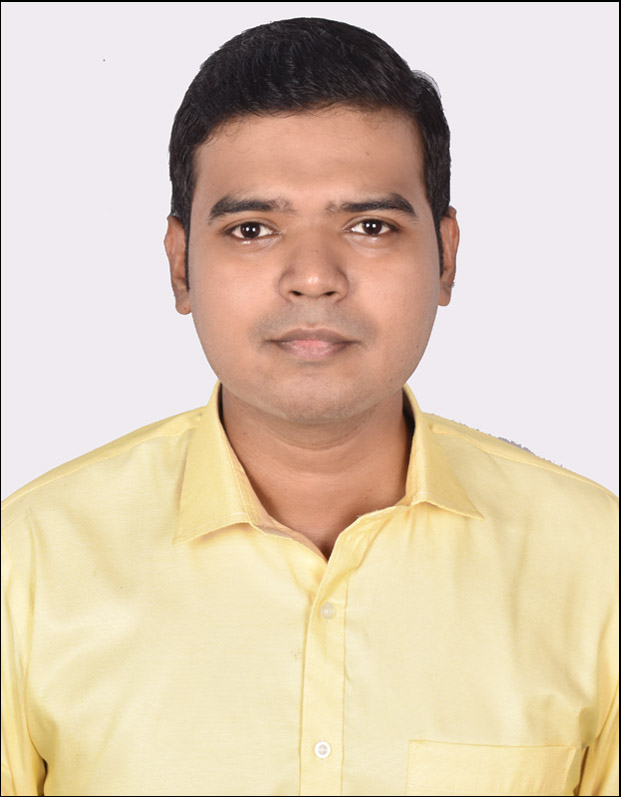}}]{Nitish Kumar} (Graduate Student Member, IEEE)
received the bachelor’s degree in electronics and communication engineering from the Sri Siddhartha Institute of Technology, Tumkur, India, in 2019. He is currently pursuing the master’s degree in communication and signal processing with the Department of Electrical Engineering, Indian Institute of Technology Hyderabad (IITH), Sangareddy,
India. His research areas include radar signal processing,
radar–camera sensor fusion, wireless communication, advanced driver assistance systems (ADAS), and model predictive control (MPC).
\end{IEEEbiography}

\vspace{11pt}
\vspace{-33pt}
\begin{IEEEbiography}
[{\includegraphics[width=1in,height=1.25in,clip,keepaspectratio]{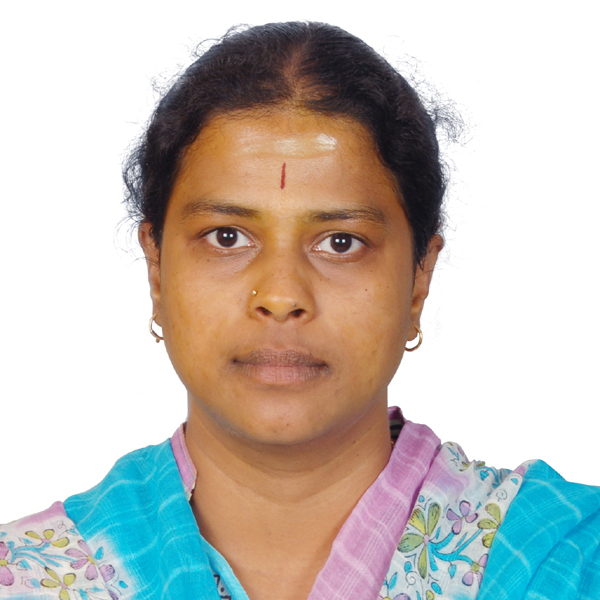}}]{Rajalakshmi Pachamuthu} (Senior Member, IEEE)
received the Ph.D. degree from Indian Institute of
Technology (IIT) Madras, Chennai, India, in 2008.
She is currently a Professor with Indian Insti-
tute of Technology Hyderabad, Sangareddy, India.
With more than 100 internationally reputed research
publications, her research interests include the Inter-
net of Things, artificial intelligence, cyber-physical
systems, sensor networks, wireless networks, drone-
based sensing, and light detection and ranging
(LiDAR)-based traffic sensing
\end{IEEEbiography}

\vfill


\begin{thebibliography}{1}
\bibliographystyle{IEEEtran}

\bibitem{ref139} H. Guo, C. Shen, H. Zhang, H. Chen, and R. Jia, “Simultaneous
trajectory planning and tracking using an MPC method for cyber-
physical systems: A case study of obstacle avoidance for an intelligent
vehicle,” IEEE Trans. Ind. Informat., vol. 14, no. 9, pp. 4273–4283,
Sep. 2018.

\bibitem{ref140} P. Hang, X. Xia, G. Chen, and X. Chen, “Active safety control of automated electric vehicles at driving limits: A tube-based MPC approach,” IEEE Trans. Transport. Electrific., vol. 8, no. 1, pp. 1338–1349, Mar. 2022.

\bibitem{ref138} C.-L. Hwang, C.-C. Yang, and J. Y. Hung, “Path tracking of an autonomous ground vehicle with different payloads by hierarchical improved fuzzy dynamic sliding-mode control,” IEEE Trans. Fuzzy Syst., vol. 26, no. 2, pp. 899–914, Apr. 2018.

\bibitem{l1}A. Thakur, C. A. R. Ram, S. Sathvik and P. Rajalakshmi, "LiDAR-GNSS Fusion to Initiate Localization at Intermediate Points on a 3D Point Cloud Map," 2024 IEEE 100th Vehicular Technology Conference (VTC2024-Fall), Washington, DC, USA, 2024, pp. 1-5, doi: 10.1109/VTC2024-Fall63153.2024.10757568.

\bibitem{l2}A. Thakur, C. A. R. Ram, S. Sathvik, B. Badugu, S. Shinde and P. Rajalakshmi, "Autonomous Cooperative Platooning Powered by LiDAR-Guided Adaptive Cruise Control," 2024 IEEE 99th Vehicular Technology Conference (VTC2024-Spring), Singapore, Singapore, 2024, pp. 1-5, doi: 10.1109/VTC2024-Spring62846.2024.10683147.

\bibitem{ref141} G. Chen, M. Hua, W. Liu, J. Wang, S. Song, and C. Liu, “Planning and tracking control of full drive-by-wire electric vehicles in unstructured scenario,” 2023, arXiv:2301.02753.

\bibitem{2_39}N.H. Amer, et al., "Modelling and control strategies in path tracking control for autonomous ground vehicles: A review of state of the art and challenges," Int. J. Robot. Syst., vol. 8(6), pp. 225–254, 2017.

\bibitem{ref_2020}F. Mohseni, E. Frisk and L. Nielsen, "Distributed Cooperative MPC for Autonomous Driving in Different Traffic Scenarios," in IEEE Transactions on Intelligent Vehicles, vol. 6, no. 2, pp. 299-309, June 2021, doi: 10.1109/TIV.2020.3025484.

\bibitem{anil2023trajectory}A. Anil and V. R. Jisha, "Trajectory Tracking Control of an Autonomous Vehicle using Model Predictive Control and PID Controller," 2023 International Conference on Control, Communication and Computing (ICCC), Thiruvananthapuram, India, 2023, pp. 1-6, doi: 10.1109/ICCC57789.2023.10164867.

\bibitem{li2022design}J. Li, K. Wu, Y. Jiang and F. Meng, "Design of Autonomous Vehicle Trajectory Tracking System Based on MPC in Unknown Environment," 2022 41st Chinese Control Conference (CCC), Hefei, China, 2022, pp. 3812-3819, doi: 10.23919/CCC55666.2022.9902834.

\bibitem{mekala2020speed}G. K. Mekala, N. R. Sarugari and A. Chavan, "Speed Control in Longitudinal Plane of Autonomous Vehicle Using MPC," 2020 IEEE International Conference for Innovation in Technology (INOCON), Bangluru, India, 2020, pp. 1-5, doi: 10.1109/INOCON50539.2020.9298213.

\bibitem{kim2023reinforcement}Y. Kim, D. -S. Pae, S. -H. Jang, S. -W. Kang and M. -T. Lim, "Reinforcement Learning for Autonomous Vehicle using MPC in Highway Situation," 2022 International Conference on Electronics, Information, and Communication (ICEIC), Jeju, Korea, Republic of, 2022, pp. 1-4, doi: 10.1109/ICEIC54506.2022.9748810.

\bibitem{yang2024collision}H. Yang, Y. He, Y. Xu and H. Zhao, "Collision Avoidance for Autonomous Vehicles Based on MPC With Adaptive APF," in IEEE Transactions on Intelligent Vehicles, vol. 9, no. 1, pp. 1559-1570, Jan. 2024, doi: 10.1109/TIV.2023.3337417.

\bibitem{mohseni2021distributed}F. Mohseni, E. Frisk and L. Nielsen, "Distributed Cooperative MPC for Autonomous Driving in Different Traffic Scenarios," in IEEE Transactions on Intelligent Vehicles, vol. 6, no. 2, pp. 299-309, June 2021, doi: 10.1109/TIV.2020.3025484.

\bibitem{yoon2021interaction}Y. Yoon, C. Kim, J. Lee and K. Yi, "Interaction-Aware Probabilistic Trajectory Prediction of Cut-In Vehicles Using Gaussian Process for Proactive Control of Autonomous Vehicles," in IEEE Access, vol. 9, pp. 63440-63455, 2021, doi: 10.1109/ACCESS.2021.3075677.

\bibitem{li2024fast}S. Li et al., "Fast Online Computation of MPC-based Integrated Decision Control for Autonomous Vehicles," in IEEE Transactions on Intelligent Vehicles, doi: 10.1109/TIV.2024.3393013.

\bibitem{li2024distributed}H. Li, T. Zhang, S. Zheng and C. Sun, "Distributed MPC for Multi-Vehicle Cooperative Control Considering the Surrounding Vehicle Personality," in IEEE Transactions on Intelligent Transportation Systems, vol. 25, no. 3, pp. 2814-2826, March 2024, doi: 10.1109/TITS.2023.3253878.

\bibitem{tsolakis2024model}A. Tsolakis, R. R. Negenborn, V. Reppa and L. Ferranti, "Model Predictive Trajectory Optimization and Control for Autonomous Surface Vessels Considering Traffic Rules," in IEEE Transactions on Intelligent Transportation Systems, vol. 25, no. 8, pp. 9895-9908, Aug. 2024, doi: 10.1109/TITS.2024.3357284

\bibitem{lpv1}A. Morsi, H. S. Abbas, S. M. Ahmed and A. M. Mohamed, "Model Predictive Control Based on Linear Parameter-Varying Models of Active Magnetic Bearing Systems," in IEEE Access, vol. 9, pp. 23633-23647, 2021, doi: 10.1109/ACCESS.2021.3056323.

\bibitem{lpv2}C. Jia, J. Cui, W. Qiao and L. Qu, "Real-Time Model Predictive Control for Battery-Supercapacitor Hybrid Energy Storage Systems Using Linear Parameter-Varying Models," in IEEE Journal of Emerging and Selected Topics in Power Electronics, vol. 11, no. 1, pp. 251-263, Feb. 2023, doi: 10.1109/JESTPE.2021.3130795.

\bibitem{lpv3}A. Elkamel, A. Morsi, M. Darwish, H. S. Abbas and M. H. Amin, "Model Predictive Control of Linear Parameter-Varying Systems Using Gaussian Processes," 2022 26th International Conference on System Theory, Control and Computing (ICSTCC), Sinaia, Romania, 2022, pp. 452-457, doi: 10.1109/ICSTCC55426.2022.9931885.


\bibitem{mpcc1}Y. Yoon, C. Kim, J. Lee and K. Yi, "Interaction-Aware Probabilistic Trajectory Prediction of Cut-In Vehicles Using Gaussian Process for Proactive Control of Autonomous Vehicles," in IEEE Access, vol. 9, pp. 63440-63455, 2021, doi: 10.1109/ACCESS.2021.3075677.

\bibitem{mpcc2}J. Li, K. Wu, Y. Jiang and F. Meng, "Design of Autonomous Vehicle Trajectory Tracking System Based on MPC in Unknown Environment," 2022 41st Chinese Control Conference (CCC), Hefei, China, 2022, pp. 3812-3819, doi: 10.23919/CCC55666.2022.9902834.

\bibitem{mpcc3}Y. Kim, D. -S. Pae, S. -H. Jang, S. -W. Kang and M. -T. Lim, "Reinforcement Learning for Autonomous Vehicle using MPC in Highway Situation," 2022 International Conference on Electronics, Information, and Communication (ICEIC), Jeju, Korea, Republic of, 2022, pp. 1-4, doi: 10.1109/ICEIC54506.2022.9748810.

\bibitem{mpcc4}H. Yang, Y. He, Y. Xu and H. Zhao, "Collision Avoidance for Autonomous Vehicles Based on MPC With Adaptive APF," in IEEE Transactions on Intelligent Vehicles, vol. 9, no. 1, pp. 1559-1570, Jan. 2024, doi: 10.1109/TIV.2023.3337417.

\bibitem{mpcc5}H. Li, T. Zhang, S. Zheng and C. Sun, "Distributed MPC for Multi-Vehicle Cooperative Control Considering the Surrounding Vehicle Personality," in IEEE Transactions on Intelligent Transportation Systems, vol. 25, no. 3, pp. 2814-2826, March 2024, doi: 10.1109/TITS.2023.3253878.


\bibitem{mpcc6}S. Li et al., "Fast Online Computation of MPC-based Integrated Decision Control for Autonomous Vehicles," in IEEE Transactions on Intelligent Vehicles, doi: 10.1109/TIV.2024.3393013.



\bibitem{mpcc7}Funke, J., et al.: Collision Avoidance and Stabilization for Autonomous Vehicles in Emergency Scenarios. IEEE Trans. Control Syst. Technol. 25(4), 1204–1216 (2017).

\bibitem{mpcc8}A. Artuñedo, M. Moreno-Gonzalez, and J. Villagra, "Lateral control for autonomous vehicles: A comparative evaluation," Annual Reviews in Control, vol. 57, article 100910, 2024. ISSN 1367-5788.

\bibitem{mpcc9}M. Rokonuzzaman, N. Mohajer, and S. Nahavandi, "NMPC-based controller for autonomous vehicles considering handling performance," in Proc. IEEE 7th International Conference on Control, Mechatronics and Automation, Delft, 2019.

\bibitem{mpcc10}L. Tang, et al., "An improved kinematic model predictive control for high-speed path tracking of autonomous vehicles," IEEE Access, vol. 8, pp. 51400-51413, 2020.

\bibitem{mpcc11}K.A. Rajagopalan, "Slip-aware model predictive optimal control for path following," in IEEE International Conference on Robotics and Automation (ICRA), Stockholm, 2016.

\bibitem{mpcc12} Y. Chen, Y. Song, L. Shi, and J. Gao, “Stochastic model predictive control for driver assistance control of intelligent vehicles considering uncertain driving environment,” J. Vib. Control, vol. 29, nos. 3–4, 2022, Art. no. 10775463211052353.

\bibitem{2_154}Kuhne, F., Lages, W.F., Silva, J.M.G.d.: Point stabilization of mobile robots with nonlinear model predictive control. In: IEEE International Conference on Mechatronics and Automation, Niagara Falls (2005)

\bibitem{2_162}A. Katriniok, et al., "Optimal vehicle dynamics control for combined longitudinal and lateral autonomous vehicle guidance," in Proc. European Control Conference (ECC), Zurich, 2013.

\bibitem{IET}M. Rokonuzzaman, N. Mohajer, S. Nahavandi, and S. Mohamed, "Review and performance evaluation of path tracking controllers of autonomous vehicles," IET Intelligent Transport Systems, vol. 15, no. 5, pp. 646-670, 2021. ISSN 1751-956X.


\bibitem{rev2}W. Liu et al., "A Systematic Survey of Control Techniques and Applications in Connected and Automated Vehicles," in IEEE Internet of Things Journal, vol. 10, no. 24, pp. 21892-21916, 15 Dec.15, 2023, doi: 10.1109/JIOT.2023.3307002. 

\bibitem{cas1}J. Andersson, J. Åkesson and M. Diehl, "Dynamic optimization with CasADi," 2012 IEEE 51st IEEE Conference on Decision and Control (CDC), Maui, HI, USA, 2012, pp. 681-686, doi: 10.1109/CDC.2012.6426534.

\bibitem{cas2}M. Fevre, P. M. Wensing and J. P. Schmiedeler, "Rapid Bipedal Gait Optimization in CasADi," 2020 IEEE/RSJ International Conference on Intelligent Robots and Systems (IROS), Las Vegas, NV, USA, 2020, pp. 3672-3678, doi: 10.1109/IROS45743.2020.9341586.

\bibitem{testbed}Pawar, D.S., Singh, A., Rajalakshmi, P. (2023). Connected Autonomous Vehicles (CAV) Testbed at IIT Hyderabad. In: Rastogi, R., Bharath, G., Singh, D. (eds) Recent Trends in Transportation Infrastructure, Volume 1. TIPCE 2022. Lecture Notes in Civil Engineering, vol 354. Springer, Singapore. 

\bibitem{tihan}
Technology Innovation Hub on Autonomous Navigation and Data Acqui-
sition Systems, "TiHAN, IIT Hyderabad"\\
\href{https://tihan.iith.ac.in/}{https://tihan.iith.ac.in/}

\bibitem{curv1}F. B. Romero, I. E. Soriano, R. Marotta and L. De Matteis, "Optimizing Motion Planning for Autonomous Vehicles: Path Planning, Curvature Analysis, and Steering Control," 2024 International Symposium on Power Electronics, Electrical Drives, Automation and Motion (SPEEDAM), Napoli, Italy, 2024, pp. 569-573, doi: 10.1109/SPEEDAM61530.2024.10609199.

\bibitem{curv2}Nengjian Wang, Defu Zhang, Lijie Zhou and Qinhui Liu, "Near optimal path planning for vehicle with heading and curvature constraints," 2010 8th World Congress on Intelligent Control and Automation, Jinan, 2010, pp. 4514-4519, doi: 10.1109/WCICA.2010.5554096.

\bibitem{curv3}S. Upadhyay and A. Ratnoo, "Continuous-Curvature Path Planning With Obstacle Avoidance Using Four Parameter Logistic Curves," in IEEE Robotics and Automation Letters, vol. 1, no. 2, pp. 609-616, July 2016, doi: 10.1109/LRA.2016.2521165. 

\bibitem{curv4}W. Yao, N. Qi, C. Yue and N. Wan, "Curvature-Bounded Lengthening and Shortening for Restricted Vehicle Path Planning," in IEEE Transactions on Automation Science and Engineering, vol. 17, no. 1, pp. 15-28, Jan. 2020, doi: 10.1109/TASE.2019.2916855. 


\end{thebibliography}
\end{document}